# Bayes Nets in Educational Assessment: Where the Numbers Come From

**Robert J. Mislevy, Russell G. Almond[*], Duanli Yan, and Linda S. Steinberg**
Educational Testing Service
Princeton, NJ 08541

## Abstract

As observations and student models become complex, educational assessments that exploit advances in technology and cognitive psychology can outstrip familiar testing models and analytic methods. Within the Portal conceptual framework for assessment design, Bayesian inference networks (BINs) record beliefs about students' knowledge and skills, in light of what they say and do. Joining evidence model BIN fragments—which contain observable variables and pointers to student model variables—to the student model allows one to update belief about knowledge and skills as observations arrive. Markov Chain Monte Carlo (MCMC) techniques can estimate the required conditional probabilities from empirical data, supplemented by expert judgment or substantive theory. Details for the special cases of item response theory (IRT) and multivariate latent class modeling are given, with a numerical example of the latter.

## 1 INTRODUCTION

This paper arises from research about assessment design from the perspective of evidentiary reasoning, as developed in Schum (1994). It focuses on statistical methods for managing uncertainty about students' knowledge, as evidenced by their performances and productions in assessment tasks. Previous work discusing cognitive psychological issues includes (Mislevy, 1995, Steinberg & Gitomer, 1996); probability-based reasoning (Almond et al., 1999; Mislevy & Gitomer, 1996); assessment design (Almond & Mislevy, in press; Mislevy, Steinberg, & Almond, in press); and computer-based simulation (Mislevy et al., in press; Steinberg & Gitomer, 1996).

Section 2 sketches a conceptual framework for assessment design that sets the stage for the building blocks of the statistical model. They are *student model*

[*] Corresponding author: ralmond@ets.org.

Bayesian inference network (SM-BIN) fragments which contain unobservable variables that characterize aspects of students' knowledge or skills, and conforming *evidence model* fragments (EM-BINs) which contain observable variables and pointers to parent student-model variables. The BIN fragments can be assembled on the fly to update belief about students' proficiencies as evidence arrives, an example of "knowledge based model construction" (KBMC; Breese, Goldman, & Wellman, 1994).

Section 3 addresses the perennial question, "Where do the numbers come from?" We describe a probability model and a Bayesian approach for estimating the parameters of student and evidence models, calibrating new tasks into an existing assessment, and drawing inferences about students. Section 4 plays these ideas out for computerized adaptive testing (CAT) with item response theory (IRT) models. Section 5 provides a numerical example of another special case, a multivariate latent class model.

## 2 THE ASSESSMENT FRAMEWORK

The essential problem is drawing inferences about what a student knows or can do, from limited observations of what she actually says or does. Our ongoing project on evidence-centered assessment design ("Portal") has two facets: developing (1) the conceptual framework for an assessment and (2) processes for developing that framework for a specific model. Figure 1 is a schematic representation of four high-levels objects in a Portal conceptual assessment framework.

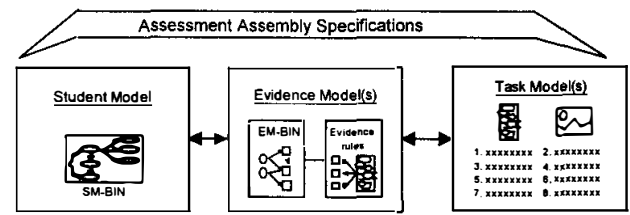

Figure 1: High-Level Assessment Design Objects

- The *Student Model* contains unobservable variables, denoted $\theta_i = (\theta_{i1},\dots,\theta_{iK})$ for Examinee $i$, that characterize the aspects of knowledge and skill that are the targets of inference. The SM-BIN manages our uncertain knowledge about $\theta_i$. The student model variables for all examinees is denoted $\theta$.

- An *Evidence Model* first describes how to extract the key items of evidence from what a student says or does in the context of a task (the work product). The evidence rules produce the values of observable variables, denoted $X_j = (X_{j1},\dots,X_{jM})$ for Task $j$. An evidence model also describes, in terms of the form for an EM-BIN, the structure by which $X_j$ depends on $\theta$. The complete collection of responses across all examinees and all tasks is denoted $\mathbf{X}$.

- A *Task Model* describes the features of a task that need to be specified. This includes specifications for the work environment, the work products, stimulus materials, and interactions between the examinee and the task, as consistent with the evidentiary requirements of a conformable evidence model. The salient characteristics of a task are expressed by task model variables, $Y_j = (Y_{j1},\dots,Y_{jL})$ for Task $j$. The complete collection of task features for all tasks in the pool is denoted $\mathbf{Y}$.

- The *Assembly Model* describes the mixture of tasks that go into an operational assessment, either the specification of a fixed test form or a procedure for determining tasks dynamically.

# 3   THE PROBABILITY FRAMEWORK

In assessment design, scientific knowledge about the domain identifies the nature of the targeted knowledge and skill, the ways in which aspects of that knowledge are evidenced in performance, and the features of situations that provide an opportunity to observe those behaviors. We encode that knowledge a student model and a series of evidence models. The key conditional independence assumption posits that the aspects of proficiency expressed in the student model account for the associations among responses to different tasks (although we may allow for conditional dependence among multiple responses within the same task).

## 3.1   THE PROBABILITY MODEL

The pertinent variables in assessment obviously include tasks' $Y$s, all of which are observable; examinees' $\theta$s, which are not; and potentially observable $X$s. Structures and parameters that reflect interrelationships among these variables, consistent with our knowledge about them, are also needed. We may build the required structures by starting with SM-BINs and EM-BINs. Sections 4 and 5 illustrate these general structures with examples.

The SM-BIN for Examinee $i$ is a probability distribution for $\theta_i$. An assumption of exchangeability posits a common prior distribution for all examinees before any responses are observed, with beliefs about expected levels and associations among components expressed through the structure of the model and higher-level parameters $\lambda$; whence, for all Examinees $i$,

$$\theta_i \sim p(\theta|\lambda). \qquad (1)$$

Depending on the strength with which theory and experience inform population-level beliefs, $p(\lambda)$ could range from vague to precise.

As noted above, the *evidence model* for a class of tasks contains rules for ascertaining the values of observable variables $X$ and a probability model for $X$ given $\theta$. We focus attention on the latter. Evidence models, indexed by the $s$, each support a class of tasks that provide values for a similar set of observable variables $X_{(s)}$; further, the structure of dependence of these $X_{(s)}$ on $\theta$ is identical for all tasks $j$ using the same evidence model. Thus the EM-BINS for task sharing the same evidence model will have the same graphical structure and exchangeable parameters (probability tables) although the conditional probability distributions within that structure can differ. As illustrated in Sections 4 and 5, this structure is guided by theory about proficiency in the domain and careful task construction that evokes targeted aspects of that proficiency.

Let $\pi_{(s)j}$ denote the parameters of the EM-BIN distributions of Task $j$ which uses the structure of evidence model $s(j)$ (or more simply, $s$). The distribution of the responses of Examinee $i$ to Task $j$, which follows evidence model $s$, is thus

$$X_{(s)ij} \sim p(X_{(s)}|\theta_i, \pi_{(s)j}). \qquad (2)$$

All the tasks using an Evidence Model $s$ produce observables $X_{(s)}$ in the same forms, furnishing information about the same components of $\theta$. Within this common evidentiary structure, however, features of the tasks can vary in ways that moderate these relationships. For example, unfamiliar vocabulary and complex sentence structures tend to make reading comprehension tasks more difficult. The parameters $\pi_{(s)j}$ for particular tasks may thus be modeled as exchangeable within evidence models given the values of designated task model variables $Y_{(s)}$; that is,

$$\pi_{(s)j} \sim p(\pi_{(s)}|Y_{(s)j}, \eta_s) \qquad (3)$$

again with prior beliefs expressed through higher-level distributions $p(\eta_s)$. We are assuming that $X_{(s)ij}$ does not depend on $Y_{(s)j}$ other than possibly through $\pi_{(s)j}$. The complete collection of probabilities for all EM-BINs for all tasks is denoted $\pi$, and the complete collection of a prior parameters for those probabilities is denoted $\eta$.

The full probability model for the responses $X_{(s)ij}$ of $N$ examinees to $J$ tasks nested within $S$ evidence models can now be written as

$$p(\mathbf{X}, \theta, \pi, \eta, \lambda | \mathbf{Y}) \sim$$

$$\prod_s \prod_j \prod_i p\left(X_{(s)ij} | \theta_i, \pi_{(s)j}\right) p\left(\pi_{(s)j} | Y_{(s)j}, \eta_s\right)$$

$$\times p(\eta_s) p(\theta_i | \lambda) p(\lambda) \qquad (4)$$

Figure 2 is a generalized form of an acyclic directed graph ("DAG") representation of this model, with boxes representing repeated elements of the same kind (Spiegelhalter et al., 1996). The structure and the nature of the distributions is tailored to the particulars of an application. In the sequel, we omit the evidence model subscripts (s) from $Xj$, $Yj$, and $\pi j$ when they are not needed.

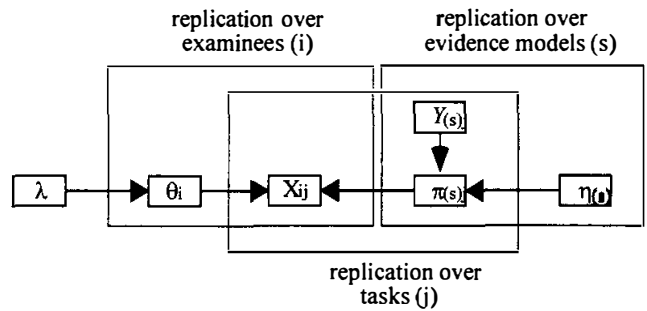

Figure 2: DAG representation of the Probability Model. $X_{ij}$ is the response of Student $i$ to Task $j$; $\theta_i$ is the parameter of Examinee $i$; $\lambda$ is the parameter of the distribution of $\theta$s; $\pi_{(s)j}$ is the parameter for Task $j$, which uses Evidence Model $s$; $Y_{(s)j}$ are the task model variables for Task $j$; and $\eta_{(s)}$ is the parameter of the distribution of $\pi_{(s)j}$s. All of these parameters can be vector-valued.

## 3.2    STATISTICAL INFERENCE

In general, the second step of Bayesian inference involves conditioning on observed data. Following from the preceding section, this would mean conditioning on whatever observations $X$ are made (say $\mathbf{X}_{old}$), to yield a posterior distribution for the unobservable parameters $\theta, \pi, \eta$, and $\lambda$, and predictive distributions for $X$s not yet observed (say $\mathbf{X}_{new}$); i.e., $p(\mathbf{X}_{new}, \theta, \pi, \eta, \lambda | \mathbf{Y}, \mathbf{X}_{old})$. Parameters and unobserved responses that are not of immediate interest can be integrated out of this joint posterior to provide marginal posterior distributions for specified variables, as required for the job at hand.

What jobs are at hand in an operational assessment? Primarily, we want to learn about the $\theta$s of individual examinees, for purposes such as making selection decisions, planning instruction, providing feedback on learning, informing policy-makers, and guiding students' work in a coached practice system. Usually we can observe a student's responses to only a limited number of tasks. On the other hand, we can often observe responses to a given task from a large number of examinees. From these observations we refine our knowledge about about how responses to a given task depend on $\theta$. This knowledge is used for selecting tasks to administer to examinees, updating our beliefs about their $\theta$s, and estimating the conditional probability distributions for new items.

### 3.2.1    Inference about Examinees

Consider inference about Examinee $i$ when $\pi, \eta$, and $\lambda$ are known to take the values of $\pi^*, \eta^*$, and $\lambda^*$ respectively. This situation may be approximated in an ongoing program with considerable data about these higher-level parameters. Suppose we observe Examinee $i$'s possibly vector-valued responses to tasks that we denote for convenience 1 through $J$. The objective is to proceed from the prior distribution $p(\theta_i | \lambda^*)$ to the posterior

$$p(\theta_i | \lambda^*, x_{i1}, \ldots, x_{iJ}, \pi_1^*, \ldots, \pi_J^*).$$

The SM-BIN for Examinee $i$ is a probability distribution for $\theta_i$. Its initial status is $p(\theta_i | \lambda^*)$. Following (2), the EM-BIN for Task 1 is $p(X_1 | \theta_i, \pi_1^*)$. Together they imply the joint distribution of $\theta_i$ and $X_1$, namely $p(X_1, \theta_i | \lambda^*, \pi_1^*) = p(X_1 | \theta_i, \pi_1^*) p(\theta_i | \lambda^*)$. Once $x_{i1}$ is observed, Bayes Theorem yields an updated distribution for $\theta_i$: $p(\theta_i | \lambda^*, x_{i1}, \pi_1^*)$. To it we can attach the EM-BIN for Task 2, $p(X_2 | \theta_i, \pi_2^*)$, and use Bayes Theorem again to obtain $p(\theta_i | \lambda^*, x_{i1}, x_{i2}, \pi_1^*, \pi_2^*)$ once $x_{i2}$ is observed. So on through Task $J$. Note that the capability to dock evidence-model BIN fragments with the student-model BIN fragment, absorb evidence from it, then discard it in preparation for the next task, is made possible by the conditional independence structure across observations from different tasks—a structure generally achieved only through careful study of proficiency in the domain and principled task construction in its light.

When both the student model variables and the observable variables are all discrete, the belief updating equations all have closed form. Complications arise when one wishes to assemble fragments on the fly, in ensuring that a proper join tree can be secured for each concatentated BIN. Almond et al. (1999) offer one solution to this problem: forcing edges in the student-model BIN among student-model variables which are parents of some observable in any evidence model that may be used.

Rarely are $\pi, \eta$, and $\lambda$ known with certainty. Fully Bayesian inference deals with them and all the $\theta$s at once (Section 3.2.2). The modularity of SM-BINs and EM-BINs that suits KBMC can be maintained by using facsimiles that replace $\pi^*$ and $\lambda^*$ with point estimates $\hat{\pi}$ and $\hat{\lambda}$—e.g., posterior means given $\mathbf{X}_{old}$—or marginal approximations $\hat{p}(\theta) = \int p(\theta|\lambda,\mathbf{X}_{old})p(\lambda)d\lambda$ and

$\hat{p}(X_{(s)ij}|\theta_i) = \int \int p\left(\pi_{(s)j}|\mathbf{X}_{old}, Y_{(s)j}, \eta_s\right)p\left(\eta_s\right)d\pi_{(s)j}d\eta_s$.

### 3.2.2    Inference about Higher-level Parameters

At the initiation of an operational assessment program, one may obtain responses from a sufficiently large sample of examinees to draw sharp inferences about the parameters of the population of examinees and a startup set of tasks. The inferential targets are $\lambda$, $\eta$, and $\pi_{old}$, and the relevant posterior distribution is $p(\pi_{old}, \eta, \lambda|\mathbf{Y}, \mathbf{X}_{old})$. The results of this analysis can be used to construct SM- and EM-BINs for use with future examinees.

The details of such analyses have been worked out for special cases of familiar assessment practices, such as the IRT methodologies outlined in Section 4. Recent work with Monte Carlo Markov Chain (MCMC) estimation (e.g., Gelman et al., 1995) provides a general approach that can be applied flexibly with new models as well, and suits the modular construction of probability distributions that characterizes KBMC. A full treatment of MCMC methods is beyond the current presentation. It suffices here to state the essential idea: to produce draws from a series of distributions constructed in a manner sketched below, which is equivalent in the limit to drawing from the posterior distribution of interest.

We address $p(\Theta, \pi_{old}, \eta, \lambda|\mathbf{Y}, \mathbf{X}_{old})$ in the present problem using a Gibbs sampler. Iteration $t+1$ starts with values for each of the parameters, say $\left\{\Theta^t, \pi_{old}^t, \eta^t, \lambda^t\right\}$. A value is then drawn from the following conditional distributions:

Draw $\Theta^{t+1}$ from $p\left(\Theta|\pi_{old}^t, \eta^t, \lambda^t, \mathbf{Y}, \mathbf{X}_{old}\right)$;

draw $\pi_{old}^{t+1}$ from $p\left(\pi_{old}|\Theta^{t+1}, \eta^t, \lambda^t, \mathbf{Y}, \mathbf{X}_{old}\right)$;

draw $\eta^{t+1}$ from $p\left(\eta|\Theta^{t+1}, \pi_{old}^{t+1}, \lambda^t, \mathbf{Y}, \mathbf{X}_{old}\right)$; and

draw $\lambda^{t+1}$ from $p\left(\lambda|\Theta^{t+1}, \pi_{old}^{t+1}, \eta^{t+1}, \mathbf{Y}, \mathbf{X}_{old}\right)$.

After convergence, the distribution of a large number of draws for a given parameter approximates its marginal distribution. Summaries such as posterior means and variances can be calculated; e.g., to construct self-contained SM- and EM-BIN fragments. We used the Spiegelhalter et al. (1996) BUGS program to do these things in the example in Section 5. See Gelman et al. (1995) on assessing convergence and criticizing model fit.

### 3.2.3    Inference about New Tasks

It is typical in operational assessment programs to continually add new tasks to the collection, whether for security purposes, to extend the range of ways to collect evidence, or simply to provide variety for students. We assume that the new items are created in accordance with existing task models and conformable evidence models.

In these cases we have occasion to estimate the parameters for the EM-BINs of these new tasks.

Suppose we have already obtained responses $\mathbf{X}_{old}$ from a sample of examinees for a set of tasks $1...J$, and by methods such as those described above obtained posterior distributions $p(\lambda|\mathbf{X}_{old})$, $p(\eta_{(s)}|\mathbf{X}_{old})$ for s=1...S, and $p(\pi_j|Y_j, \mathbf{X}_{old})$ for j=1...J. We wish to calibrate into the set a new Task J+1, which uses Evidence Model s[J +1] and is characterized by task features $Y_{J+1}$. We obtain responses $\mathbf{X}_{new}$ from a sample of $N_{new}$ examinees to both Task J+1 and previously-calibrated tasks. The objective now is to obtain an approximation $p(\pi_{J+1}|Y_{J+1}, \mathbf{X}_{old}, \mathbf{X}_{new})$ we can use to produce the EM-BIN for Task J+1.

A first approach acknowledges the remaining uncertainty about the parameters of the old tasks and the examinee and task hyperdistributions. Posterior distributions from the startup estimation are employed as the priors for $\lambda$, $\eta$, and $\pi_{old}$. These are, respectively, $p(\lambda|\mathbf{X}_{old})$, $p(\eta|\mathbf{X}_{old})$ and $p(\pi_{old}|\mathbf{Y}_{old}, \mathbf{X}_{old})$. The iterations in an MCMC solution echo those of the startup estimation; one draws successively for $\lambda$, $\eta$, and $\pi_{old}$ as well as for $\pi_{J+1}$ and $\Theta_{new}$. In addition to posteriors for $\pi_{J+1}$ and $\Theta_{new}$ based on $\mathbf{X}_{new}$, one obtains updated distributions for $\lambda$, $\eta$, and $\pi_{old}$ based on both $\mathbf{X}_{old}$ and $\mathbf{X}_{new}$.

A simpler approach treats the previous point estimates as known. The probability model for this so-called "empirical Bayes" approximation is

$$p\left(\mathbf{X}_{new}, \Theta_{new}, \pi_{J+1}|\hat{\lambda}, \hat{\eta}, \hat{\pi}_1, ..., \hat{\pi}_J, Y_{J+1}\right)$$

$$= p\left(\mathbf{X}_{new}|\Theta_{new}, \hat{\pi}_1, ..., \hat{\pi}_J, \pi_{J+1}\right)$$

$$\times p\left(\pi_{J+1}|\hat{\eta}_{s[J+1]}, Y_{J+1}\right)p\left(\Theta_{new}|\hat{\lambda}\right)$$

$$= \prod_{i=1}^{N_{new}} \prod_{j=1}^{J} p\left(X_{ij}|\theta_i, \hat{\pi}_j\right)p\left(X_{i,J+1}|\theta_i, \pi_{J+1}\right)$$

$$\times p\left(\pi_{J+1}|\hat{\eta}_{s[J+1]}, Y_{J+1}\right)p\left(\theta_i|\hat{\lambda}\right)$$

MCMC estimation approximates the posterior

$$p\left(\Theta_{new}, \pi_{J+1}|\mathbf{X}_{new}, \hat{\lambda}, \hat{\eta}, \hat{\pi}_1, ..., \hat{\pi}_J, Y_{J+1}\right)$$

with iterations of the following form:

Draw $\Theta^{t+1}$ from $p\left(\Theta|\pi_{J+1}^t, \hat{\lambda}, \hat{\eta}, \hat{\pi}_1, ..., \hat{\pi}_J, Y_{J+1}\right)$

Draw $\pi_{J+1}^{t+1}$ from $p\left(\pi_{J+1}|\Theta^{t+1}, \hat{\lambda}, \hat{\eta}, \hat{\pi}_1, ..., \hat{\pi}_J, Y_{J+1}\right)$

## 4  ITEM RESPONSE THEORY AND ADAPTIVE TESTING

This section discusses a special case in which the preceding ideas have been applied in large-scale operational testing programs as the Graduate Record Examination (GRE) and the Armed Services Vocational Aptitude Battery (ASVAB): computerized adaptive testing (CAT), using item response theory (IRT). Both the student model and the observations are fairly simple, and the methodologies they use have evolved over the past fifty years in the context of educational testing.

### 4.1  ITEM RESPONSE THEORY (IRT)

An IRT model expresses an examinee's propensity to perform well in a domain of test items, in terms of a single unobservable proficiency variable $\theta$. Item responses are posited to be independent, conditional on $\theta$ and item parameters that express characteristics such as items' difficulty or sensitivity to proficiency. The Rasch model for $J$ dichotomous test items is a simple example:

$$P(x_1,\ldots,x_J \mid \theta,\beta_1,\ldots,\beta_J) = \prod_{j=1}^{J} P(x_j \mid \theta,\beta_j) \qquad (5)$$

where $x_j$ is the response to Item $j$ (1 for right, 0 for wrong), $\beta_j$ is the difficulty parameter of Item $j$, and $P(x_j \mid \theta,\beta_j) = \exp[x_j(\theta-\beta_j)]/[1+\exp(\theta-\beta_j)]$. The $\beta_j$s play the role of the $\pi_j$s in the notation of Section 3.

The student model in IRT contains the single proficiency variable $\theta$, and an SM-BIN is a probability distribution for $\theta$—initially $p(\theta)$. A task model specifies a set of salient features of a class of items, regarding work products, cognitive demands, item format, and so on, as may be required to assemble tests or to model item parameters. An evidence model contains the rules for determining the value of the response $x_j$ from an examinee's work product, be it a rubric for a human to evaluate a free response or engineering specs to parse marks on a photosensitive answer sheet. An evidence model also specifies the structure of EM-BINs, which in this example are identical in form but generally differ as to the value of $\beta_j$. The evidence model may further posit a relationship between $\beta_j$ and $Y_j$ (see Section 4.3).

The likelihood function (5) takes the form of catenated EM-BINs. Once an examinee's response vector $x = (x_1,\ldots,x_J)$ is observed, it is viewed as a likelihood function for $\theta$, say $L(\theta \mid x,\mathbf{B})$. Bayesian inference is based on the posterior $p(\theta \mid x,\mathbf{B}) \propto L(\theta \mid x,\mathbf{B})p(\theta)$, where $\mathbf{B} = (\beta_1,\ldots,\beta_J)$. $p(\theta \mid x,\mathbf{B})$ can be summarized by its posterior mean $\overline{\theta}$ and variance $Var(\theta \mid x,\mathbf{B})$.

### 4.2  INFERENCE ABOUT EXAMINEES: CAT

A fixed test form provides different accuracy for different values of $\theta$, with greater precision when $\theta$ lies in the neighborhood of the items' difficulties. CAT tailors the test's level of difficulty to each examinee. Testing proceeds sequentially, with each successive item $k+1$ selected to be informative about the examinee's $\theta$ in light of the responses to the first $k$ items, or $\mathbf{x}^{(k)}$ (Wainer et al., 1990, Chap 5). A Bayesian approach to CAT starts from a prior distribution for $\theta$ and selects each next item to minimize expected posterior variance, or $E_{x_j}\left[Var\left(\theta \mid x^{(k)}, x_j, \mathbf{B}^{(k)}, \beta_j\right) \mid x^{(k)}, \mathbf{B}^{(k)}\right]$. Additional constraints on item selection can be incorporated into the assessment assembly algorithm, such as item content and format encoded as task model variables $Y_j$. Testing ends when a desired measurement accuracy has been attained or a specified number of items has been presented.

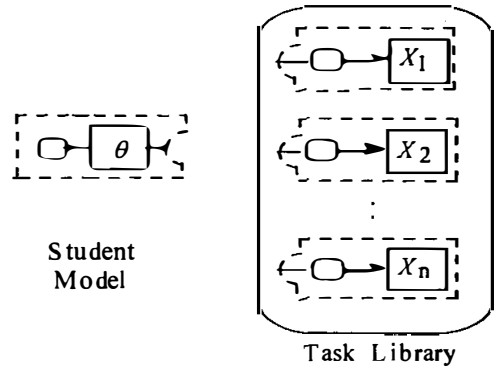

**Student Model**

**Task Library**

a) S M -B I N and Task/E M -B I N Library

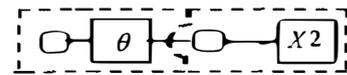

b) E M -B I N for Item 2 "docked" with S M -B I N

Figure 3: SM-BIN and Task/EM-BINs in IRT-CAT. The distribution object for the SM-BIN contains the distribution for $\theta$; those for the tasks contain the conditional distributions of the item response given $\theta$.

Figure 3 depicts the SM-BIN and EM-BINs in IRT-CAT. Figure 3a shows the SM-BIN on the left, consisting of the single SM variable $\theta$ and the distribution object that contains current belief about its unknown value. On the right is a library of EM-BINs, each linked to a particular task. The observable variable $x_j$ appears, along with the distribution object that contains the IRT conditional distribution for $x_j$ given $\theta$. Figure 3b shows an EM-BIN "docked" with the SM-BIN to absorb evidence in the form of a response to the corresponding item.

### 4.3    INFERENCE ABOUT HIGHER-LEVEL PARAMETERS

For selecting items and scoring examinees in typical applications, estimates of the item parameters are obtained from large samples of examinee responses and treated as known. This procedure plays the role of the MCMC estimation described in Section 3.2.2. Bayes modal estimation and maximum likelihood (Bock & Aitkin, 1981) are widely used, although MCMC methods are appearing (e.g., Albert, 1992).

There is growing interest in exploiting collateral information about test items' features $Y_j$ to reduce the number of pretest examinees needed to estimate item parameters (Mislevy, Sheehan, & Wingersky, 1993). For example, Scheuneman, Gerritz, and Embretson (1991) accounted for about 65-percent of the variance in item difficulties in the Reading section of the National Teacher Examination with variables for tasks' syntactic complexity, semantic content, cognitive demand, and knowledge demand. Fischer (1973) integrated cognitive information into IRT by modeling Rasch item difficulty parameters as linear functions of effects for item features. Incorporating a residual term to allow for variation of difficulties among items with the same features gives

$$\beta_j = \sum_{k=1}^{K} Y_{kj} \eta_k + \varepsilon_j,$$

where $\eta_k$ is the contribution of Feature $k$ to the difficulty of an item, $Y_{kj}$ is the extent to which Feature $k$ is represented in Item $j$; and $\varepsilon_j$ is a N(0,$\varphi^2$) residual term.

### 4.4    INFERENCE ABOUT NEW TASKS

CAT selects items according to their difficulty parameters in order to maximize information about an examinee's $\theta$. To do this one must know something about the $\beta_j$s. Now testing programs continually introduce new items into the item pool so items do not become spuriously easy because examinees share them. Estimating the $\beta$s of new items within the operational testing context is called "online calibration." This is usually done by administering examinees both optimally-determined items whose $\beta$s are well-estimated and randomly-selected new items whose $\beta$s are not known. The responses to the former are used to determine the examinee's operational score, while the responses to the latter are used to learn about the new items' $\beta$s. This is the situation discussed in Section 3.2.3. It is standard practice to estimate the parameters of new items using the empirical Bayes approximation; i.e., the parameters of the "old" items are treated as known. Empirical studies have shown this expedient yields satisfactory estimates for $\mathbf{B}_{new}$. The evidentiary value of $Y$s for $\beta$s can also be exploited in online calibration, in order to reduce the number of pretest examinees that are needed.

Analogous procedures are necessary for operational assessments in the multivariate framework. Section 5.6 addresses this need.

## 5    A LATENT CLASS MODEL

This section concerns binary skills latent class models (Haertel, 1984). We give numerical results obtained in analyses of data from Tatsuoka's (1990) research on mixed number subtraction with middle school students.

### 5.1    BINARY SKILLS MODELS

In a binary skills model, the student model contains a vector of $K$ 0/1 variables $\theta_i = (\theta_{i1}, \ldots, \theta_{iK})$, each of which signifies that an examinee either does (1) or does not (0) possess some particular element of skill or knowledge in some learning domain. A task in this domain is similarly characterized by a vector of $K$ 0/1 task model variables $Y_j = (Y_{j1}, \ldots, Y_{jK})$ that indicates whether a task does (1) or does not (0) require each of these skills for successful solution. The statistical component of the evidence model posits that an examinee is likely to succeed on a task ($X_j = 1$) when she possesses the skills it demands, and likely to fail ($X_j = 0$) if she lacks one or more of them.

### 5.2    THE METHOD B NETWORK

This example is grounded in a cognitive analysis of middle-school students' solutions of mixed-number subtraction problems. Klein et al. (1981) identified two methods of solution:

*Method A*: Convert mixed numbers to improper fractions, subtract, then reduce if necessary.

*Method B*: Separate mixed numbers into whole number and fractional parts, subtract as two subproblems, borrowing one from minuend whole number if necessary, then simplify and reduce if necessary.

We focus on students learning to use Method B. The cognitive analysis mapped out a flowchart for applying Method B to a universe of fraction subtraction problems. A number of key procedures appear, which a given problem may or may not require in accordance with its structure. Students had trouble solving a problem with Method B when they could not carry out one or more of the procedures an item required. Instruction was designed to review them. A student model based on five procedures that are sufficient for mixed-number subtraction problems when no common denominator needs to be found is thus suited to planning review sessions for a student. The procedures are as follows:

Skill 1:  Basic fraction subtraction.

Skill 2:  Simplify/reduce fraction or mixed number.

Skill 3:  Separate whole number from fraction.

Skill 4:  Borrow one from the whole number in a given mixed number.

Skill 5: Convert a whole number to a fraction.

$\theta_1, \ldots, \theta_5$ are student-model variables that correspond to having or not having each of these skills. Prior analyses revealed that Skill 3 is a prerequisite to Skill 4. We introduced a three-level variable, $\theta_{WN}^{'}$, that incorporates this constraint. Level 0 is having neither of these skills; Level 1 is having Skill 3 but not Skill 4; Level 2 is having both of them.

Table 1 lists fifteen items from Dr. Tatsuoka's data set, characterized by the skills they require—i.e., their $Y$s. The list is grouped by patterns of skill requirements. All the items in a group have the same structural relationship to $\theta$. They require a student have the same conjunction of skills in order to make a "true positive" correct response.

Table 1: Skill Requirements for Fractions Items

| ITEM | TEXT | SKILLS REQUIRED 1 | 2 | 3 | 4 | 5 | EM |
|---|---|---|---|---|---|---|---|
| 1 | $\frac{6}{7} - \frac{4}{7} =$ | x | | | | | 1 |
| 2 | $\frac{3}{4} - \frac{3}{4} =$ | x | | | | | 1 |
| 3 | $\frac{11}{8} - \frac{1}{8} =$ | x | x | | | | 2 |
| 4 | $3\frac{1}{2} - 3\frac{1}{2} =$ | x | | x | | | 3 |
| 5 | $4\frac{4}{5} - 1\frac{4}{5} =$ | x | | x | | | 3 |
| 6 | $3\frac{7}{8} - 2 =$ | x | | x | | | 3 |
| 7 | $3\frac{1}{2} - 2\frac{1}{2} =$ | x | | x | x | | 4 |
| 8 | $4\frac{1}{2} - 2\frac{4}{5} =$ | x | | x | x | | 4 |
| 9 | $7\frac{3}{5} - \frac{4}{5} =$ | x | | x | x | | 4 |
| 10 | $4\frac{1}{3} - 1\frac{5}{3} =$ | x | | x | x | | 4 |
| 11 | $4\frac{1}{10} - 2\frac{8}{10} =$ | x | | x | x | | 4 |
| 12 | $2 - \frac{1}{3} =$ | x | | x | x | x | 5 |
| 13 | $3 - 2\frac{1}{5} =$ | x | | x | x | x | 5 |
| 14 | $7 - 1\frac{4}{3} =$ | x | | x | x | x | 5 |
| 15 | $4\frac{4}{12} - 2\frac{7}{12} =$ | x | x | x | x | | 6 |

We will re-analyze data that Dr. Tatsuoka collected and analyzed using her Rule-Space methodology, which also used a binary skills foundation but with a somewhat different set of skills and a pattern-matching approach to handling uncertainty. We will consider the responses of 325 students judged to be using Method B.

### 5.3   THE PROBABILITY MODEL

The full probability distribution for all 325 examinees and 15 items has the form shown in (4). The distributions are specified as follows.

The student model variables are $(\theta_1, \ldots, \theta_5, \theta_{WN})$ Preliminary analyses based on point estimates from Tatsuoka's analysis led us to the structure depicted in Figure 4. Edges represent conditional dependence relationships, with directions chosen according to the usual teaching order. Recalling that each of the variables

$\theta_1, \ldots, \theta_5$ is binary and $\theta_{WN}$ has three levels, we may describe the SM-BIN, or $p(\theta|\lambda)$, as follows:

$\theta_1$ is Bernoulli with probability $\lambda_1$; i.e., $\theta_1 \sim \text{Bern}(\lambda_1)$.

$\theta_2$ depends on $\theta_1$: $\theta_2|\theta_1 = z \sim \text{Bern}(\lambda_{2z})$ for $z=0,1$.

$\theta_5$ depends on $\theta_1$ and $\theta_2$: $\theta_5|(\theta_1 + \theta_2 = z) \sim \text{Bern}(\lambda_{5z})$ for $z=0,1,2$.

$\theta_{WN}$ can take values 0,1,2; it depends on $\theta_1$, $\theta_2$, and $\theta_5$: $\theta_{WN}|(\theta_1 + \theta_2 + \theta_5 = z) \sim \text{Cat}(\lambda_{WN,z,0}, \lambda_{WN,z,1}, \lambda_{WN,z,2})$, for $z=0,1,2,3$.

$\theta_3 = 0$ if $\theta_{WN} = 0$; $\theta_3 = 1$ if $\theta_{WN} = 1$ or 2.

$\theta_4 = 0$ if $\theta_{WN} = 0$ or 1; $\theta_4 = 1$ if $\theta_{WN} = 2$.

The last two of these relationships are logical rather than probabilistic.

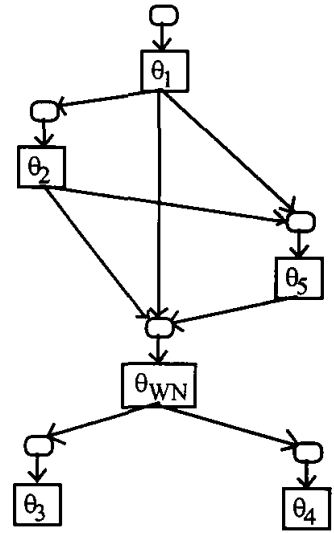

Figure 4: DAG for student model for mixed number subtraction.    Squares represent student-model variables; roundtangles represent distribution objects.

We specified, for each $\lambda$, a prior distribution with an effective sample size of 25. These are $\text{Beta}(\alpha, \beta)$ for the $\lambda$ s that are parameters of Bernoulli distributions, with $\alpha = 21$ and $\beta = 6$ when the probability is expected to be high (e.g., students who have Skill 1 are likely to have Skill 2) and vice versa when the probabilities are expected to be low (students who don't have Skill 1 probably don't have Skill 2 either). We used Dirichlet priors for the $\lambda_5$ vectors, positing increasing belief of having Skills 3 and 4 as a student has more of Skills 1, 2, and 5.

Evidence models correspond to patterns of $\theta_1, \ldots, \theta_5$ that are required to solve a class of items, at least one of which appears in the 15 item data set. There are six of them, equivalently characterized by the vector of skills required or by the pattern of Task Model variables $Y$ of items that conform with that evidence model. The evidence models and the items that use them can be read from Table 1. For example, Evidence Model 3 is characterized by $Y = (1, 0, 1, 0, 0)$, and Items 4-6 use this model.

The EM-BINs take the form of misclassification matrices, specified by a false positive probability $\pi_{j0}$ of a correct response if the examinee does not have the conjunction of skills for the evidence model Task $j$ uses, and a true positive probability $\pi_{j1}$ of a correct response if she does. We denote by $\delta_{i(s)}$ whether Examinee $i$ has the skills needed for tasks using evidence model $s$ ; it takes the value 1 if she does and 0 if she does not.

The EM-BIN for Task $j$, which uses evidence model $s$, contains the observable response $X_{ij}$, pointers to the student model variables for which $Y_{(s)k} = 1$, and the following conditional probability distributions:

$$X_{ij}\big|\big(\delta_{i(s)} = z\big) \sim \text{Bern}\big(\pi_{jz}\big), \text{ for } z = 0, 1.$$

These probabilities are allowed to differ from item to item, both within and across evidence models. Figure 5 shows the structure of EM-BINs for $s = 2$ and 4.

For priors for the $\pi$s, we again imposed Beta distributions with effective sample sizes of 25. These are Beta(21,6) for $\pi_{j1}$s, or true positives, and Beta(6,21) for $\pi_{j0}$s, or false positives. This corresponds to the expectation that students who do have the necessary skills will answer an item correctly about .8 of the time, and students who don't will answer correctly only about .2 of the time. (These priors are just initial guesses. We expect and indeed observe substantial changes in the posterior means.)

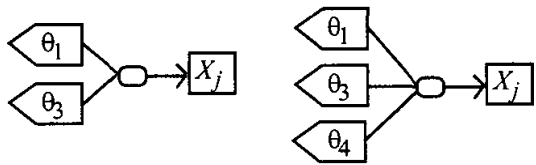

Figure 5: EM-BIN structures for tasks using Evidence Models 2 and 4. Distribution object represents distributions of response $X_j$ given values of student-model parents indicated by pointers to student-model variables.

## 5.4    INFERENCE ABOUT EXAMINEES

In an operational assessment, inference about an individual examinee starts with the possibly-diffuse population prior distribution—i.e., the SM-BIN initialized at $p\big(\theta\big|\hat{\lambda}\big)$ or at $p\big(\theta\big|X_{old}\big) = \int p\big(\theta\big|\lambda\big)p\big(\lambda\big|X_{old}\big)d\lambda$, depending on the approximation being used. EM-BINs

for the items to which responses are observed are joined with the SM-BIN, and evidence is absorbed into the SM-BIN (Mislevy, 1995).

Fixing the values of the $\lambda$ s and $\pi$s were fixed at the posterior means of the first run in the following section, standard Bayes net propagation algorithms can be used to update believes of $\theta$ (Almond, et al., 1999). Table 2 gives the prior and posterio beliefs for an examinee who gave mostly correct answers to items requiring skills other than Skill 2, but not those that do require it. The base-rate and the updated probabilities show substantial shifts toward the belief that this examinee has Skills 1, 3, 4, and possibly 5, but almost certainly not Skill 2.

Table 2: Profile for $X = (1, 1, 0, 1, 1, 0, 1, 1, 0, 1, 1, 1, 0, 1, 0)$

| SKILL | PRIOR PROB. | POSTERIOR PROB. |
|---|---|---|
| 1 | .883 | .999 |
| 2 | .618 | .056 |
| 3 | .937 | .995 |
| 4 | .406 | .702 |
| 5 | .355 | .561 |

## 5.5    INFERENCE ABOUT HIGHER-LEVEL PARAMETERS

As a baseline against which to compare subsequent runs that better mirror operational work, we used BUGS to estimate the full probability model from Section 5.3 with all 15 items and all 325 examinees. Table 3 gives summary statistics from this run for selected parameters. The posterior means and standard deviations of the parameter estimates appear, along with method-of-moments estimates of Beta distributions these posteriors imply. Recalling the priors were Beta distributions with an effective weight of 25 observations, the column labeled $\hat{n}$ approximates the effective number of observations the data was worth in estimating each parameter. They are always less than the actual sample size of 325, since examinees' actual skill vectors are not known with certainty.

Table 4 simulates a startup run in an operational assessment program. 225 of the examinees were sampled, and parameters were estimated in BUGS for the $\lambda$ s and for the $\pi$s of 12 items. This run establishes the statistical framework for subsequent inferences about new examinees and new items. The rows with values show very similar posterior means to those of the baseline run, but slightly higher standard deviations. Translated to approximate Beta distributions, they show proportionally lower effective sample sizes. The omitted rows correspond to the 3 items not administered; they are the "new" items to which we now turn our attention.

Table 3: MCMC Estimation, All Tasks, 325 Examinees

| Parameter/State | | Mean | SD | $\hat{\alpha}$ | $\hat{\beta}$ | $\hat{n}$ |
|---|---|---|---|---|---|---|
| $\lambda_1$ | | .81 | .02 | 204 | 49 | 226 |
| $\lambda_2$ | $\lambda_1=0$ | .21 | .07 | 11 | 23 | 8 |
| | $\lambda_1=1$ | .90 | .03 | 134 | 11 | 118 |
| $\pi_4$ | FalsePos | .19 | .05 | 12 | 51 | 37 |
| | TruePos | .92 | .02 | 193 | 16 | 182 |
| $\pi_5$ | FalsePos | .20 | .04 | 16 | 63 | 52 |
| | TruePos | .91 | .02 | 173 | 18 | 164 |
| $\pi_8$ | FalsePos | .09 | .02 | 20 | 211 | 204 |
| | TruePos | .87 | .03 | 114 | 17 | 104 |
| $\pi_{10}$ | FalsePos | .04 | .01 | 9 | 199 | 181 |
| | TruePos | .81 | .03 | 109 | 26 | 108 |
| $\pi_{12}$ | FalsePos | .18 | .03 | 38 | 169 | 180 |
| | TruePos | .75 | .04 | 109 | 36 | 118 |
| $\pi_{14}$ | FalsePos | .05 | .01 | 12 | 218 | 203 |
| | TruePos | .68 | .04 | 90 | 42 | 106 |

Table 4: MCMC Estimation, 12 Tasks, 225 Examinees

| Parameter/State | | Mean | SD | $\hat{\alpha}$ | $\hat{\beta}$ | $\hat{n}$ |
|---|---|---|---|---|---|---|
| $\lambda_1$ | | .80 | .03 | 144 | 37 | 154 |
| $\lambda_2$ | $\lambda_1=0$ | .23 | .08 | 6 | 21 | 1 |
| | $\lambda_1=1$ | .90 | .03 | 96 | 10 | 80 |
| $\pi_4$ | FalsePos | .15 | .05 | 8 | 42 | 23 |
| | TruePos | .92 | .02 | 135 | 11 | 119 |
| $\pi_8$ | FalsePos | .10 | .02 | 17 | 155 | 145 |
| | TruePos | .83 | .04 | 65 | 14 | 52 |
| $\pi_{12}$ | FalsePos | .16 | .03 | 23 | 121 | 117 |
| | TruePos | .74 | .04 | 75 | 27 | 74 |

## 5.6   INFERENCE ABOUT NEW TASKS

We carried out two BUGS runs to calibrate the three new items into the assessment. The response data for both runs are the same: responses to all 15 items from the 100 examinees not used in the setup run.

Table 5 summarizes the results from the Bayesian approximation, in which the $\lambda$ s and the $\pi$ s about which evidence was obtained in the first run are started with Beta or Dirichlet priors that reflect the posteriors from the setup run, via the method of moments approximations. For these parameters, the resulting posteriors agree well with the results from the 325-examinee setup run—they are based on the same examinees, although the responses to the three new items from the 225 startup sample of examinees is not included. The posteriors for the three new items, correspondingly, do not match quite as closely and translate to lower effective sample sizes.

Table 6 summarizes the results from the empirical Bayes approximation, in which the $\lambda$ s and the $\pi$ s about which evidence was obtained in the first run are fixed at the pos-terior means obtained in the setup run. The only parameters involved in the MCMC iterations were the 100 new examinees' $\theta$ s and the 3 new items' $\pi$ s. We see that the posterior means for the new items agree almost exactly with those of the preceding Bayesian solution. The effective sample sizes are greater by about 3 on the average, which represents the impact of treating the $\lambda$ s and the $\pi$ s from the previous run as "known" rather than "less uncertain than they were." This modest overstatement of precision would seem acceptable in practical work if it simplifies operational procedures substantially.

Table 5: 3 New Tasks, 100 Examinees, Priors from Previous Run

| Parameter/State | | Mean | SD | $\hat{\alpha}$ | $\hat{\beta}$ | $\hat{n}$ |
|---|---|---|---|---|---|---|
| $\lambda_1$ | | .81 | .02 | 205 | 49 | 226 |
| $\lambda_2$ | $\lambda_1=0$ | .22 | .08 | 11 | 21 | 5 |
| | $\lambda_1=1$ | .90 | .03 | 134 | 13 | 119 |
| $\pi_4$ | FalsePos | .19 | .05 | 11 | 47 | 31 |
| | TruePos | .94 | .02 | 192 | 12 | 177 |
| $\pi_5$ | FalsePos | .27 | .07 | 11 | 30 | 14 |
| | TruePos | .89 | .03 | 79 | 10 | 62 |
| $\pi_8$ | FalsePos | .08 | .02 | 19 | 209 | 201 |
| | TruePos | .85 | .03 | 95 | 17 | 85 |
| $\pi_{10}$ | FalsePos | .09 | .03 | 8 | 79 | 59 |
| | TruePos | .79 | .05 | 49 | 13 | 35 |
| $\pi_{12}$ | FalsePos | .17 | .03 | 35 | 173 | 181 |
| | TruePos | .75 | .04 | 110 | 38 | 121 |
| $\pi_{14}$ | FalsePos | .07 | .03 | 6 | 75 | 53 |
| | TruePos | .68 | .06 | 43 | 20 | 36 |

Table 6: 3 New Tasks, 100 Examinees, Priors fixed at Posterior Means from Previous Run

| Parameter/State | | Mean | SD | $\hat{\alpha}$ | $\hat{\beta}$ | $\hat{n}$ |
|---|---|---|---|---|---|---|
| $\pi_5$ | FalsePos | .27 | .07 | 12 | 33 | 17 |
| | TruePos | .89 | .03 | 81 | 10 | 64 |
| $\pi_{10}$ | FalsePos | .09 | .03 | 8 | 80 | 61 |
| | TruePos | .80 | .05 | 48 | 12 | 34 |
| $\pi_{14}$ | FalsePos | .07 | .03 | 6 | 78 | 57 |
| | TruePos | .68 | .06 | 46 | 21 | 41 |

# 6   NEXT STEPS

There are several fronts along which further work is needed. We are currently applying the approach illustrated in Section 5 to a simulation-based assessment of problem-solving in dental hygiene. We are also considering alternative ways of joining SM- and EM-BINs that produce approximations in the SM-BIN posteriors, trading off exactitude for flexibility in larger problems. Finally, we plan to develop templates for EM-BIN probability distributions that formally incorporate cognitively-relevant task model variables into response models (e.g., Wang, Wilson, & Adams, 1997). The most important lesson we have learned is the need for coordination across specialties in the design of complex assessments. An assessment that pushes the frontiers of psychology, technology, statistics, and a substantive domain all at once cannot succeed unless all are incorporated into a coherent design from the beginning.

## Acknowledgments

The first author's work was supported in part by the National Center for Research on Evaluation, Standards, Student Testing (CRESST), Educational Research and Development Program, cooperative agreement number R117G10027 and CFDA catalog number 84.117G, as administered by the Office of Educational Research and Improvement, U.S. Department of Education. We thank Eddie Herskovitz and Andrew Gelman for their contributions to this work and to Kikumi Tatsuoka for permission to use her data on mixed number subtraction. We gratefully acknowledge our intellectual debt to Dr. Tatsuoka, having leaned on the insights in her classroom observations, cognitive analysis, test design, and analyses.